\documentclass[runningheads]{llncs}
\usepackage{graphicx}
\usepackage{caption}
\usepackage{subcaption}
\usepackage{amssymb}
\usepackage{amsmath}
\usepackage{float}
\usepackage{tikz}

\newcommand{\comment}[1]{}
\definecolor{kr-red}{RGB}{225,38,0}
\definecolor{cw-red}{RGB}{183,2,6}
\definecolor{kr-dark}{RGB}{0,0,0}
\definecolor{cw-blue}{RGB}{29,70,213}

\title{How Do You Act? \\ An Empirical Study to Understand Behavior of Deep Reinforcement Learning Agents}
\titlerunning{How Do You Act?}
\author{Richard Meyes \and
Moritz Schneider \and
Tobias Meisen}
\authorrunning{R. Meyes et al.}
\institute{Chair of Technologies and Management of the Digital Transformation, \\ University of Wuppertal, Rainer-Gruenter-Straße 21, 42119 Wuppertal, Germany \\
\email{\{meyes, m.schneider-hk, meisen\}@uni-wuppertal.de}}
\begin{document}
\maketitle 
\begin{abstract}
The demand for more transparency of decision-making processes of deep reinforcement learning agents is greater than ever, due to their increased use in safety critical and ethically challenging domains such as autonomous driving. In this empirical study, we address this lack of transparency following an idea that is inspired by research in the field of neuroscience. We characterize the learned representations of an agent's policy network through its activation space and perform partial network ablations to compare the representations of the healthy and the intentionally damaged networks. We show that the healthy agent's behavior is characterized by a distinct correlation pattern between the network's layer activation and the performed actions during an episode and that network ablations, which cause a strong change of this pattern, lead to the agent failing its trained control task. Furthermore, the learned representation of the healthy agent is characterized by a distinct pattern in its activation space reflecting its different behavioral stages during an episode, which again, when distorted by network ablations, leads to the agent failing its trained control task. Concludingly, we argue in favor of a new perspective on artificial neural networks as objects of empirical investigations, just as biological neural systems in neuroscientific studies, paving the way towards a new standard of scientific falsifiability with respect to research on transparency and interpretability of artificial neural networks.
\keywords{Transparency, Interpretability, Explainability, Deep Reinforcement Learning, Neuroscience}
\end{abstract}

\section{Introduction}
Recent research on general-purpose artificial intelligence (AI) has seen some major breakthroughs in the past few years spurred by the advances of deep reinforcement learning (DRL) algorithms utilized in environments with sparse rewards and complete information \cite{silver2016mastering,silver2017mastering} or in complex multi-agent environments with incomplete information \cite{openai2018dota,baker2019emergent,jaderberg2019human}. However, the research path leading up to today's pinnacle of these applications is marked by a crisis of reproducibility and required intense manual trial-and-error efforts such as finding a good network initialization and subsequent hyper-parameter tuning, which can make all the difference between a working and a failing solution \cite{rlblogpost}. What complicates the problem even more is that many working solutions are interspersed with unwanted behavioral artifacts that manifest in the learned policy of agents, if the environment allows for such manifestation, e.g. in the domain of learning locomotion \cite{popov2017data}. Such artifacts are commonly caused by incentivizing an agent to solely maximize a possibly richly shaped reward without any constraints on its policy. The usual approach of training agents to maximize their cumulative reward and quantitatively evaluating them solely based on this reward or any other performance metric, such as the ELO rating in chess, raises a key question: \textbf{How can we trust an agent, if we do not understand how its behavior emerges from its internal processes and the complex interplay of its individual functional components?}

In this paper, we aim to contribute towards answering this question following a research paradigm from the field of neuroscience based on empirical studies of large and complex neural systems. Such systems have been the objects of investigation for decades starting with the influential work of Hubel and Wiesel in the 1950s \cite{hubel1959receptive}, aiming to make them transparent and interpretable with respect to how their inner processes contribute to abstract concepts like consciousness and decision-making. Specifically, we investigate the behavior of DRL agents in three different classic control environments based on the learned representations of their policy networks, aiming to find a link between these representations and different behavioral stages during the execution of the trained policy. We characterize the actor's learned representations based on its layer activation during the execution of the policy and use network ablations (cf. section \ref{sec:method_ablations}) to intentionally damage agents, evoking malfunctioning behavior to compare the representations of the fully intact and damaged networks to each other.

First, we investigate the impact of network ablations with different sizes in different layers on the agent's capability to solve its trained control task and show that the the agent exhibits a task specific robustness to these ablations depending on the size and location of the ablations. We further investigate how the activations of single units contribute to solving the control task, uncovering specific correlation patterns between these activations and the executed actions during an episode. Finally, we investigate patterns in the temporal evolvement of the actor's layer activation and find that the healthy agent's learned representation contains distinct activation states that can be directly linked to the different behavioral stages of the policy that successfully solves the control task, ultimately providing a link between the agent's behavior and its internal processes.

\section{Related Work}
Most of the recent work on transparency, interpretability and explainability of AI comes from the field of computer vision (CV), where the main focus is commonly placed on investigations of convolutional neural networks (CNNs) and the importance of specific input variables for a network's output \cite{papernot2017practical,fong2017interpretable,faust2018visualizing,su2019one,fong2019understanding}. Similar efforts are made in the field of natural language processing (NLP), where recurrent neural networks (RNNs) are investigated for their representations of linguistic properties, contextual understanding or sentiment \cite{karpathy2015visualizing,radford2017learning,bau2018identifying,madsen2019visualizing}. Typically, learned network representations are characterized via embedding methods like t-SNE \cite{maaten2008visualizing} or UMAP \cite{mcinnes2018umap} visualizing the high dimensional activation-space of neural networks to identify the role of specific network components in solving a given task \cite{liu2016towards,rauber2016visualizing,elloumi2018analyzing,dibia2019convnet,carter2019activation}. To this end, network ablations were used to study the impact of single units on a network's performance \cite{dalvi2019neurox}, aiming to decide which units can be pruned without affecting a network's discriminative power \cite{molchanov2016pruning,li2016pruning,cheney2017robustness}. Subsequently, network ablations revealed that a single unit's importance can be characterized by the magnitude of its weights \cite{dalvi2019one} and the extent to which the distribution of its incoming weights changes during training \cite{meyes2019ablation_pp,meyes2019ablation}. Additionally, it was shown that units, which are easily interpretable, are not necessarily more important than units with a less accessible interpretability \cite{morcos2018importance}. Recently, controversial insights on methods how to evaluate the similarity of learned network representations have been reported and demonstrate the early stage of current knowledge and thus, the importance and the need for more research on the topic \cite{morcos2018insights,kornblith2019similarity}. 
In general, the extensive efforts of recent research aimed to map the classification result of a supervised trained network to humanly interpretable explanations. We aim to extend these efforts towards the DRL domain, where despite some work on understanding Deep Q-Networks and interpreting their learned policies in environments with a discrete action space \cite{zahavy2016graying}, to the best of our knowledge, work on facilitating transparency and interpretability of learned representations by means of network ablations has not been conducted yet. However, in view of the fact that robust DRL and its application in real world scenarios is still a matter of current research \cite{rlblogpost,dulac2019challenges}, we argue that a better interpretability of DRL agents is of utmost importance.

\section{Study Design}
\subsection{Experimental Setup}
\begin{figure}[tb!]
    \centering
    \includegraphics[width=\textwidth]{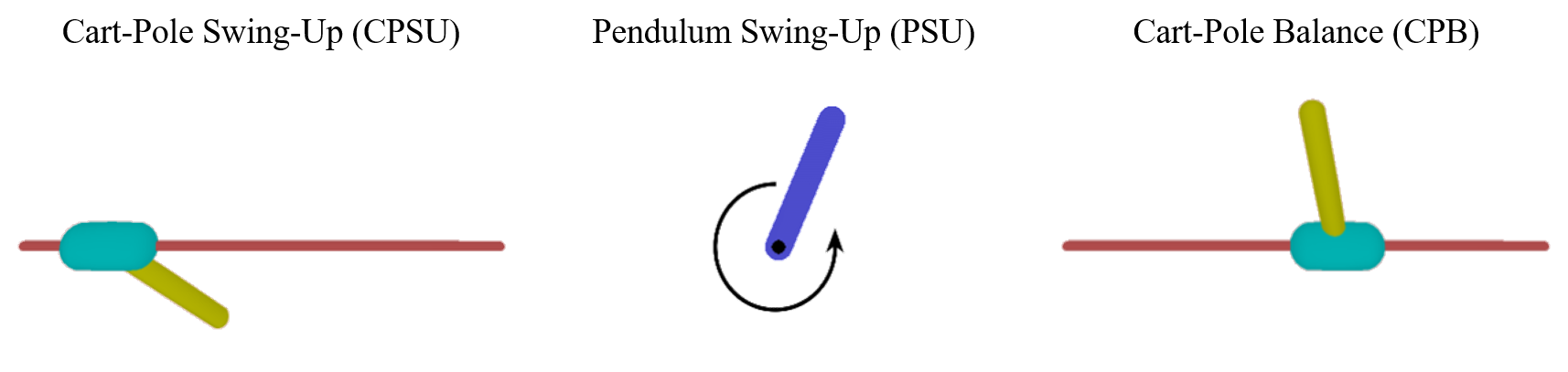}
    \caption{Three exemplary rendered images of the respective control environments.}
    \label{fig:environments}
\end{figure}

In this empirical study, we trained a DRL agent in three different classic control environments, namely the cart-pole swing-up (CPSU) environment \cite{openAIroboschool}, the pendulum swing-up (PSU) environment \cite{openAIgym} and the cart-pole balance (CPB) environment \cite{openAIroboschool} (cf. Figure \ref{fig:environments}). 
Although each environment poses an individual challenge, they share the partial objectives of controlling a cart on a rail or balancing a pendulum/pole in an upright position, providing some degree of comparability of the observed agent's behavior across tasks. We refrain from a more detailed explanation of the intricacies of these environments regarding their state space, action space and reward functions at this point, as they are well-known benchmark environments for DRL research and have been extensively explained elsewhere \cite{openAIroboschool,openAIgym}.

As the object of investigation, we trained an actor-critic agent in the three described environments with the deep deterministic policy gradient algorithm as outlined in \cite{lillicrap2015continuous}. Both, the actor and the critic network consist of two hidden layers with 400 units in the first layer and 300 units in the second layer with both layers using ReLU activation and layer normalization \cite{ba2016layer}. The critic is supplied with the actor's chosen actions, which is superimposed by an Ornstein–Uhlenbeck noise process \cite{uhlenbeck1930theory}, only in the second hidden layer. Each agent was trained for 800,000 time steps and optimized via Adam \cite{kingma2014adam} with all other hyper-parameters being the same as in \cite{lillicrap2015continuous}. All computations were performed on a single machine containing two Intel Xeon Platinum 8168 processors with a total number of 48 physical cores and 8 NVIDIA Tesla V100 32G GPUs.

\subsection{Characterization of Learned Representations}
We characterize the actor's learned representations based on its layer activation during policy execution. We use network ablations to intentionally damage the actor, evoking malfunctioning agent behavior to compare the representations of the fully intact and damaged networks. To this end, we record the activation of each single unit within the fully intact actor and its predicted actions for each time step of an episode in addition to the cumulative episodic reward to establish a baseline recording. Additionally, we record the same data for each individual ablation case to compare it to the baseline recording.

\paragraph{\textbf{Network Ablations.}} \label{sec:method_ablations} We perform partial network ablations in a single layer with varying proportions of ablated units by manually clamping their activations to zero, effectively preventing any flow of information through the ablated units. We select the amount of ablated units in a range from $5\%$ to $90\%$ in steps of $5\%$ until $30\%$ and then in steps of $10\%$ until $90\%$. In addition, we deviate from this pattern once by ablating $33.33\%$ of units within a layer. Thereby, the ablated units are selected in a sliding window manner that is shifted across the layer, similar to sliding a kernel over an image in a CNN while the window position is frozen during an episode. Note that the total number of ablations with the same proportion varies because they depend on the size of the layer, the size of the window and the stride of the window. For instance, in a layer with $300$ units and a chosen window size of $5\%$ with a stride of $10$ units, $15$ units are ablated at once resulting in 29 different network ablations in total. For all ablations, we chose a constant stride value of $10$ units to gather sufficient activation recordings for statistical analysis while at the same time keeping the computational efforts manageable.

\paragraph{\textbf{Extraction of Activation Patterns.}} \label{sec:method_patterns}
To determine how single units contribute to the control task, we calculate the Pearson correlation coefficient of its set of activations $A_{i,j}=\{a_{t}|t\in[0,T]\}$ and the outputs of the actor network $U=\{u_{t}|t\in[0,T]\}$, for each time step within an episode, where $t$ denotes the time step within the episode, $T$ denotes the total number of time steps per episode, $i$ denotes the $i$-th layer and $j$ the $j$-th unit within that layer.

Furthermore, to characterize the learned representations within a layer of the actor, we store the activations of each single unit in that specific layer for each time step of an episode in a matrix $M^{T\times N}$, where $T$ denotes the number of time steps per episode and $N$ denotes the number of units per layer. We visualize the evolvement of the actor's activation during an episode using an open source Python implementation of UMAP \cite{mcinnes2018umap} to embed the stored activations into a two-dimensional space, i.e. $M\in \mathbb{R}^{T\times2}$. Thus, each point in the embedded space represents the activation of a specific layer of the actor network for a single time step of an episode. We chose the default parameters for the UMAP embeddings after an initial attempt for finding better values for the number of nearest neighbours or the minimum distance between data points yielded no significant visual improvement of the embeddings. 

\section{Results}
\subsection{Impact of Ablations on the Agent's Capability} 
To establish a baseline evaluation, we train the healthy agent to achieve near state-of-the-art results in all three environments, i.e. a maximum total episodic reward of $886.4$ for the CPSU task, $-275.87$ for the PSU task and $1000$ for the CPB task. For reasons of performance comparability across the three environments, the absolute return is normalized so that the minimum return value in each environment is $0$ and the respective baseline return value is $1$. 

\begin{figure}[t!]
\begin{tabular}{c c c}
     & 30\% ablations in the first layer & 30\% ablations in the second layer \\
     
    \rotatebox{90}{CPSU} & \includegraphics[width=0.47\columnwidth]{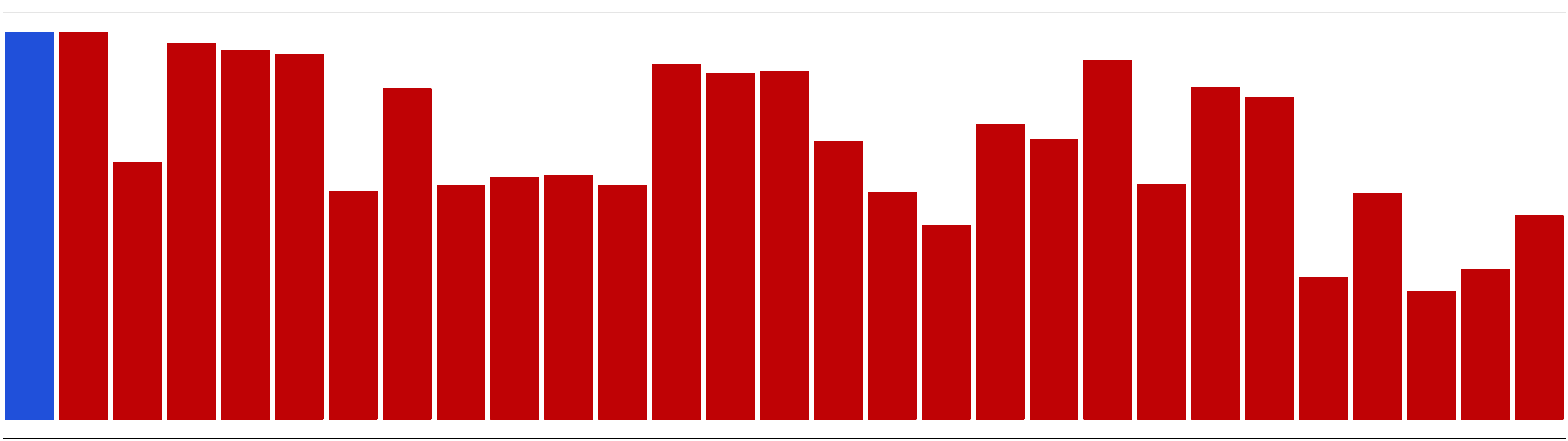} & \includegraphics[width=0.47\columnwidth]{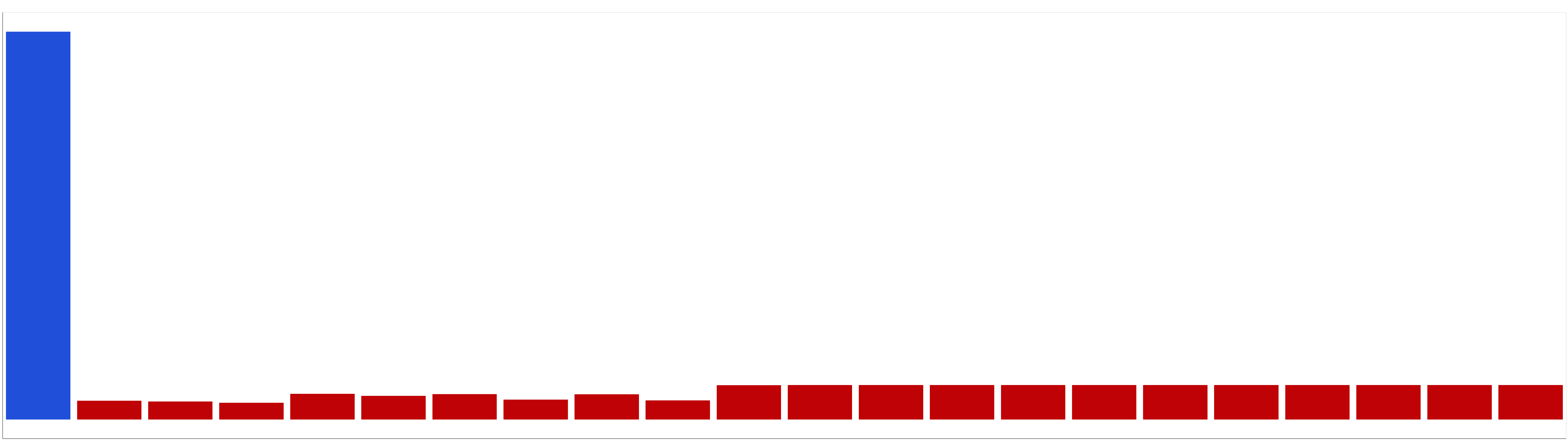} \\
    
    \rotatebox{90}{PSU} & \includegraphics[width=0.47\columnwidth]{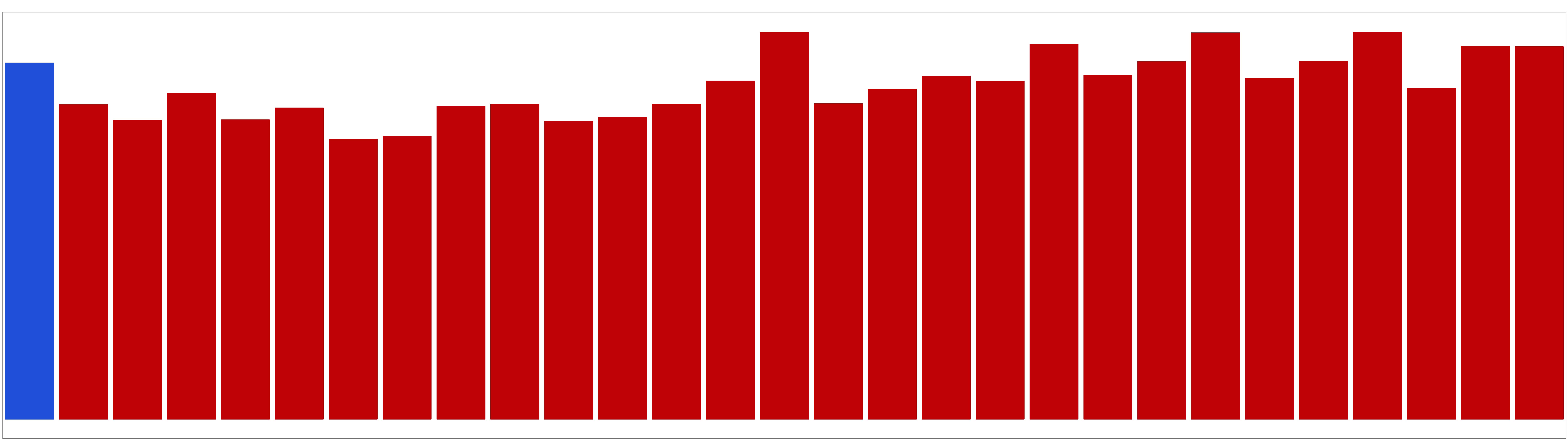} & \includegraphics[width=0.47\columnwidth]{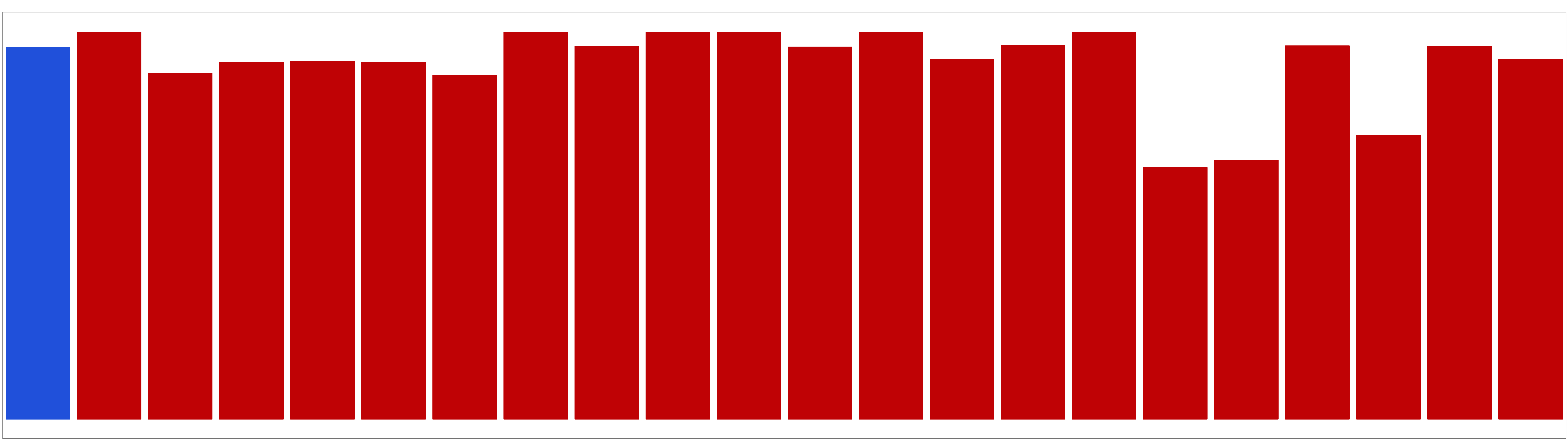} \\
    
    \rotatebox{90}{CPB} & \includegraphics[width=0.47\columnwidth]{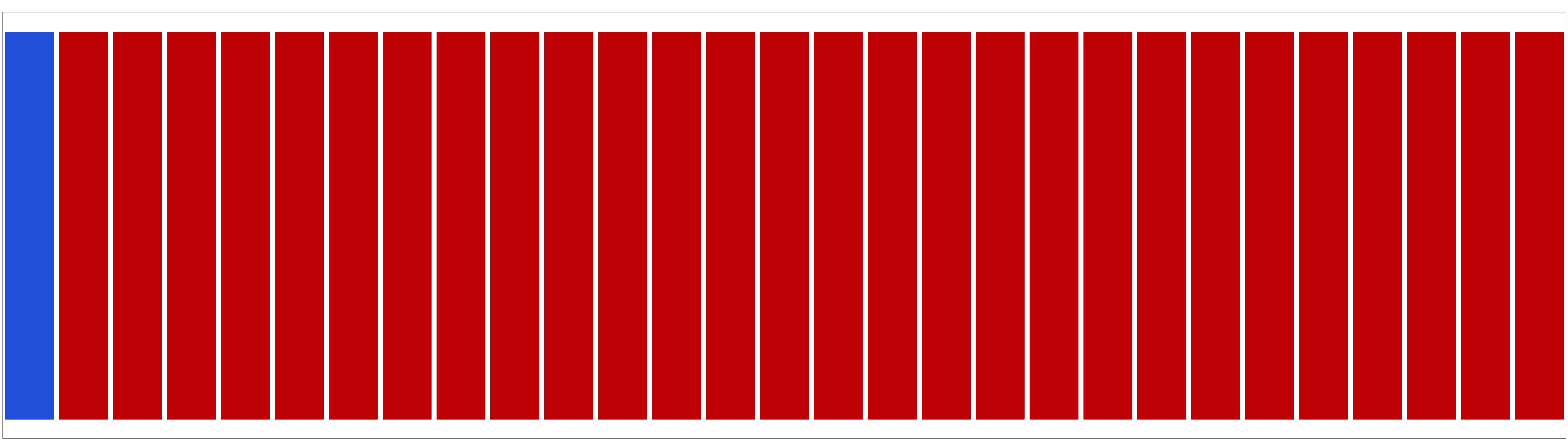} & \includegraphics[width=0.47\columnwidth]{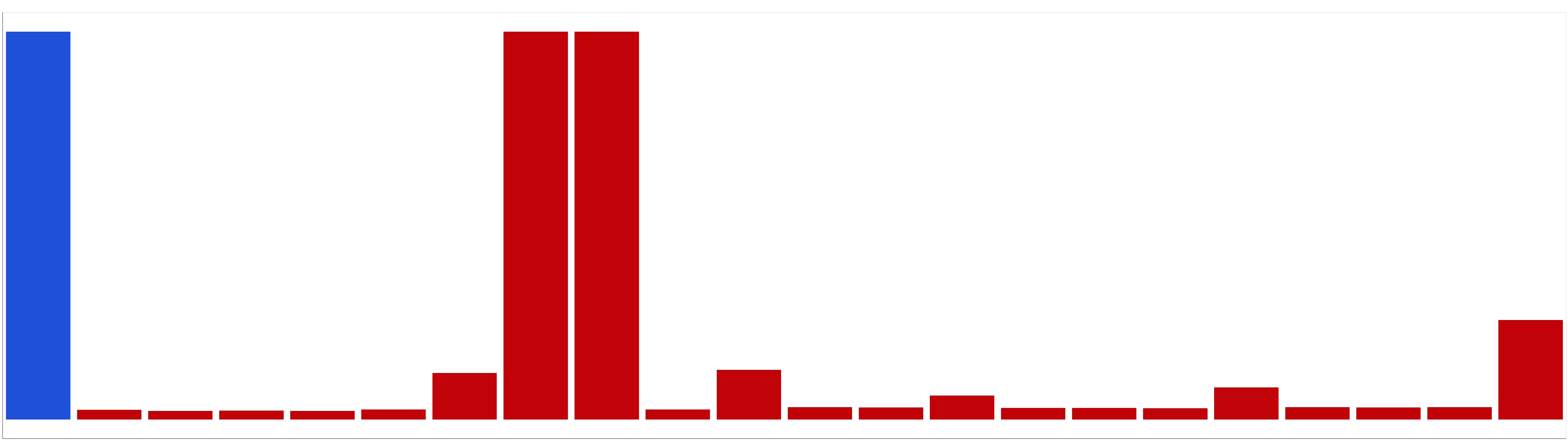} \\
\end{tabular}
\\ \vspace{0.1 cm}
\centering
\tikz\draw[cw-red,fill=cw-red] (0,0) circle (.4ex); Ablated \hspace{1.5cm}
\tikz\draw[cw-blue,fill=cw-blue] (0,0) circle (.4ex); Baseline
\caption{Comparison of the normalized returns achieved as a result of ablations of $30\%$ of the units (red bars) in to its respective baselines (blue bars).}
\label{fig:returns_barcharts_30}
\end{figure}

Figure \ref{fig:returns_barcharts_30} shows the normalized return for the baseline in comparison to all 29 network ablations in the first and second layer with a window size of $30\%$ ($120$ units) for the three control tasks. 
For both swing-up tasks, most ablations in the first layer have a negative impact on the agent's capability to solve the tasks. Interestingly, there are some ablations that have little to no impact or even a positive impact, thus increasing the return. In case of the CPB task, ablating $30\%$ of the units in the first layer does not affect the agent's capability to solve the task at all. Contrary to the first layer, all ablations in the second layer have a strong negative impact for the CPSU task and the CPB task (except for two cases), however, only a few ablations have a comparably negative impact for the PSU task, where many ablations have little to no impact or even a positive impact. The negligible impact of ablations suggests that either the capacity of the network has not been exploited to its fullest extent so that some units do not contribute to solving the task and could be pruned or that the information represented by the ablated units is redundantly represented by other units making the agent robust against network ablations. The positive impact of ablations suggests that some units may play competing roles in the learned representation and that resolving this competition by targeted ablations improves the agent's capability to solve a task. Both observations are consistent with previously reported findings on the impact of ablations in supervised trained neural networks on image recognition tasks \cite{meyes2019ablation_pp,meyes2019ablation}. 

\begin{figure}[tb!]
    \includegraphics[width=\textwidth]{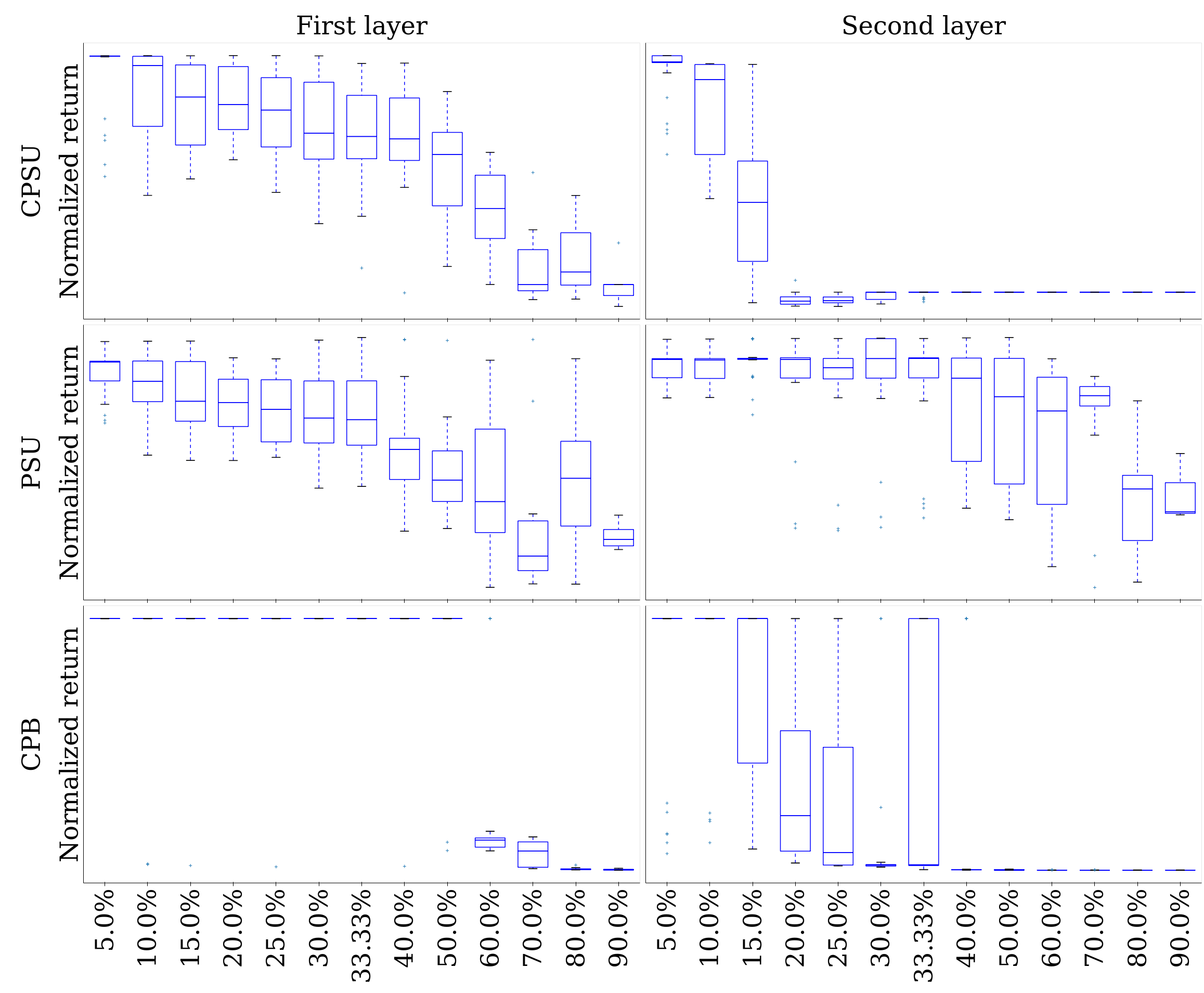}
    \caption{Distributions of the normalized returns for all ablations performed in the first layer (left side) and second layer (right side).}
    \label{fig:returns_mean_boxplot}
\end{figure}

Figure \ref{fig:returns_mean_boxplot} shows the distributions of the normalized returns resulting from the different network ablations in the first layer and second layer for the three control tasks.

On average, the return decreases proportionally to the amount of ablated units. Comparing the impacts in the first layer across the three tasks shows a similar trend for the CPSU and the PSU task, i.e. a slow but steady decrease of the achieved return with increasing sizes of ablations but a much more robust behavior for the CPB task, where ablations of up to $50\%$ generally do not affect the agent's capability to solve the task. Further, comparing the impacts in the second layer shows a similar trend for the CPSU and the CPB task, i.e. a strong and sudden decrease in the achieved return for small ablation sizes, but a much more robust behavior for the PSU task, where ablations of up to $33.33\%$ only marginally affect the agent's capability to solve the task. Interestingly, connecting the similarity of the ablation impacts with the similarity of the different tasks suggests that the first layer holds a representation of how to swing up the pole/pendulum while the second layer holds a representation of how to control the moving cart. More precisely, ablations in the first layer impact the agent in both tasks, in which a pole has to be swung up, while the representation for the task, which merely requires balancing the pole, is very robust against ablations in this layer. Analogously, ablations in the second layer strongly impact the agent in both tasks, in which a cart has to be controlled, while the representation for the task without a cart is fairly robust against ablations in this layer. These results suggest that interlinked learning objectives to solve the task such as controlling the cart, swinging up the pendulum and subsequently balancing it, are represented in different locations of the network. These observations are consistent with previously reported findings on the localized representations of specific classes in supervised trained neural networks on image classification tasks \cite{veit2016residual,olah2018the,rafegas2019understanding}.

\subsection{Impact of Ablations on Single Unit Activity (SUA)} \label{sec:results_SUA}
Following the observations described above, we wonder what role the precise interplay of SUA plays with respect to the agent's executed policy. More specifically, we ask whether the contribution of SUA to the executed actions during an episode shows a distinct pattern for the healthy agent and to what extent this pattern is distorted in case of ablations with a negative impact on the achieved return. To this end, we characterize this pattern via the set of Pearson correlation coefficients calculated for the activations of single units within a layer and the outputs of the actor network for each time step within an episode (cf. \ref{sec:method_patterns}). 

\begin{figure}[t!]
    \includegraphics[width=\textwidth]{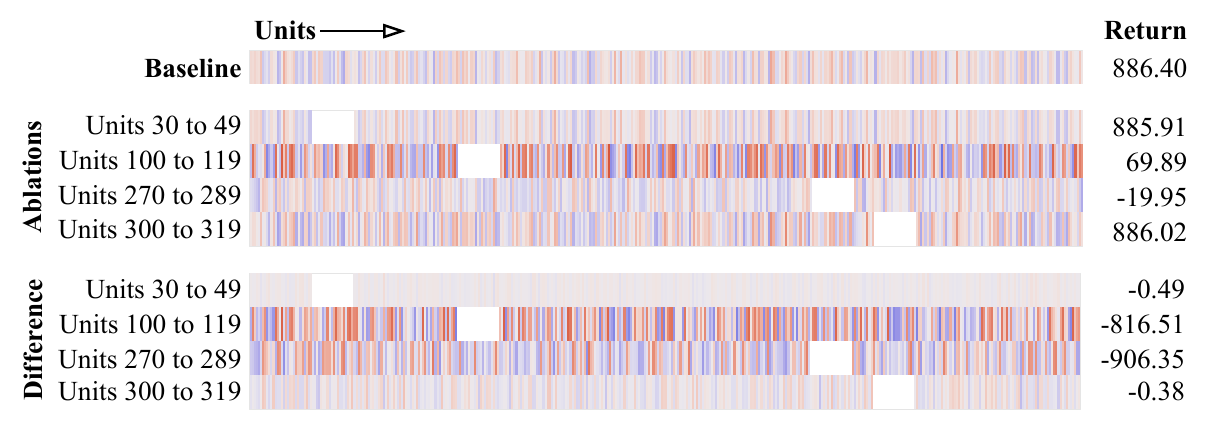}
    \centering
    \vspace{0.1 cm}
    \tikz\draw[cw-red,fill=cw-red] (0,0) circle (.4ex); Positive correlation $(\approx 1)$ \hspace{1.5cm}
    \tikz\draw[cw-blue,fill=cw-blue] (0,0) circle (.4ex); Negative correlation $(\approx -1)$
    \caption{Correlation pattern of the activations of all 400 units in the first layer during the CPSU task for the healthy agent (baseline) and four exemplary ablations, as well as the change of these patterns compared to the baseline (bottom four rows).}
    \label{fig:invertedpendulumswingup_linear1_activation_action_correlations_05_heatmap}
\end{figure}

Figure \ref{fig:invertedpendulumswingup_linear1_activation_action_correlations_05_heatmap} shows this pattern for the baseline and four exemplary ablations of $5\%$ of units in the first layer activated in the CPSU task.
Each row contains 400 entries corresponding to the 400 units in the first layer. Each entry contains the correlation value and shows how the unit's activation correlates with the actor's chosen action. The empty spaces in the rows show the ablated units, for which no correlation coefficient is calculated. The top row shows the baseline correlation pattern in comparison to the following four rows, which show the correlation patterns corresponding to the four exemplary ablations. The bottom four rows show to what extent the patterns resulting from the ablations change compared to the baseline pattern, specified by the difference between the baseline pattern and the ablation patterns. The ablations of units 100 to 119 and 270 to 289, resulting in the agent's failure to solve the task, show a general increase in correlation between the SUA and the chosen actions and the strongest difference of the pattern compared to the baseline. A high correlation value indicates a unit's exclusive contribution to a specific control direction, i.e. whenever the cart is moved to either side, specific units are selectively active and contribute to the control in a specific direction. However, such distinct contributions of single units do not seem to resemble a robust representation as we find that patterns with less distinct correlations between single unit activations and the chosen actions generally lead to higher returns. This observation shows some similarity with previously reported findings about the importance of single units in supervised trained networks for image classification tasks. Specifically, networks that memorize well instead of generalizing are more reliant on units that show a high selectivity in their activation for specific classes, indicating that units which selectively get activated for specific classes do not contribute as much to a robust and generalized representation as units with a less selective activation \cite{morcos2018importance}. 

\begin{figure}[t!]
\centering
    \tikz\draw[kr-red,fill=kr-red] (0,0) circle (.4ex); High return \hspace{1.5cm}
    \tikz\draw[kr-dark,fill=kr-dark] (0,0) circle (.4ex); Low return \hspace{1.5cm}
    \tikz\draw[cw-blue,fill=cw-blue] (0,0) circle (.4ex); Baseline
    \includegraphics[width=\textwidth]{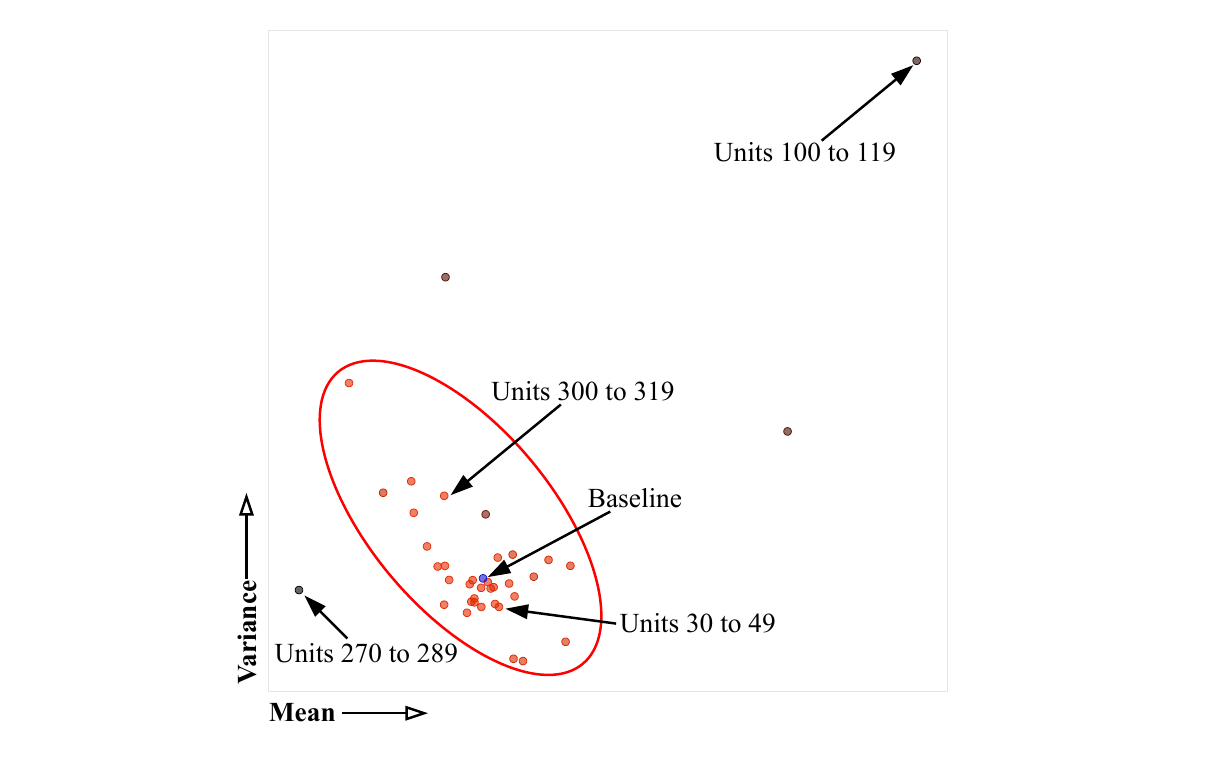}
    \caption{Scatter plot of the mean and the variance of the correlation patterns for the baseline and all 29 ablations of the size of $5\%$ and their corresponding returns in the CPSU task.}
    \label{fig:invertedpendulumswingup_linear1_activation_action_correlations_05_scatter}
\end{figure}

In order to further solidify that notion, we compared the mean and the variance of the correlation patterns of all ablations with the mean and the variance of the baseline pattern, hypothesizing that high values for the mean and the variance, corresponding to strong and distinct correlations, result in a low return. Figure \ref{fig:invertedpendulumswingup_linear1_activation_action_correlations_05_scatter} shows a scatter plot of the mean and the variance of the correlation patterns for the baseline and all 29 ablations of the size of $5\%$ and their corresponding returns.
Confirming the hypothesis, ablations of units resulting in large values for the mean and the variance, e.g. units 100 to 119 (marked in the top right corner of the scatter plot) lead to low returns. Almost all other ablations with mean and variance values close to the baseline (points within the red ellipsis) do not result in task failures but achieve returns comparable to the baseline. Interestingly, the ablation of the units 270 to 289, which results in small values for the mean and the variance, also leads to a low return, suggesting that our hypothesis can be extended towards small values for the mean and the variance, corresponding to no clear contribution for most of the single units to the control task. 

\begin{figure}[t!]
    \centering
    \tikz\draw[kr-red,fill=kr-red] (0,0) circle (.4ex); High return \hspace{1.5cm}
    \tikz\draw[kr-dark,fill=kr-dark] (0,0) circle (.4ex); Low return \hspace{1.5cm}
    \tikz\draw[cw-blue,fill=cw-blue] (0,0) circle (.4ex); Baseline
    \\ \vspace{0.1 cm}
    \begin{subfigure}{.32\textwidth}
        \centering
        \includegraphics[width=\textwidth]{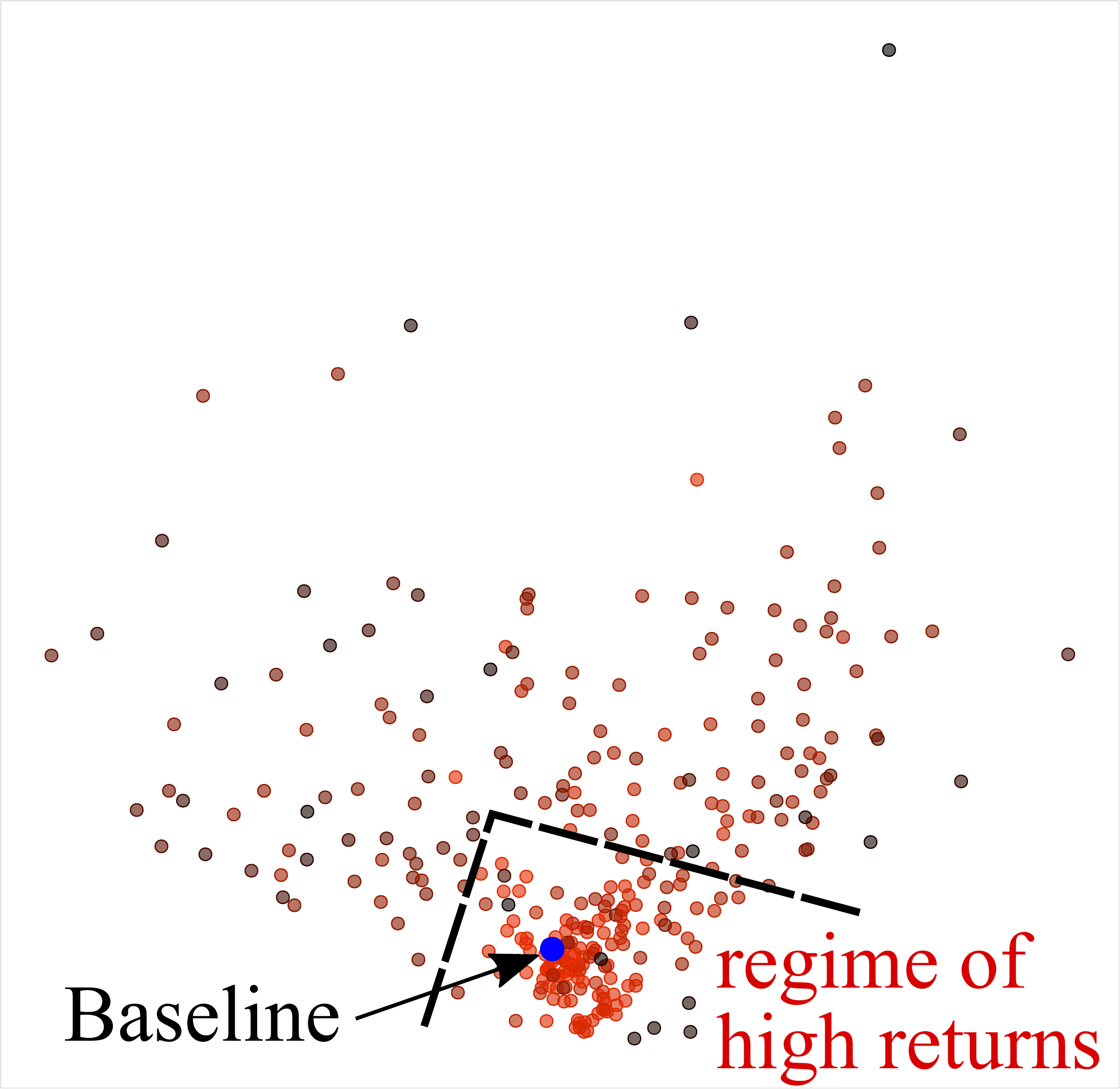}
        \subcaption{First layer, CPSU}
    \end{subfigure}
    \hfill
    \begin{subfigure}{.32\textwidth}
        \centering
        \includegraphics[width=\textwidth]{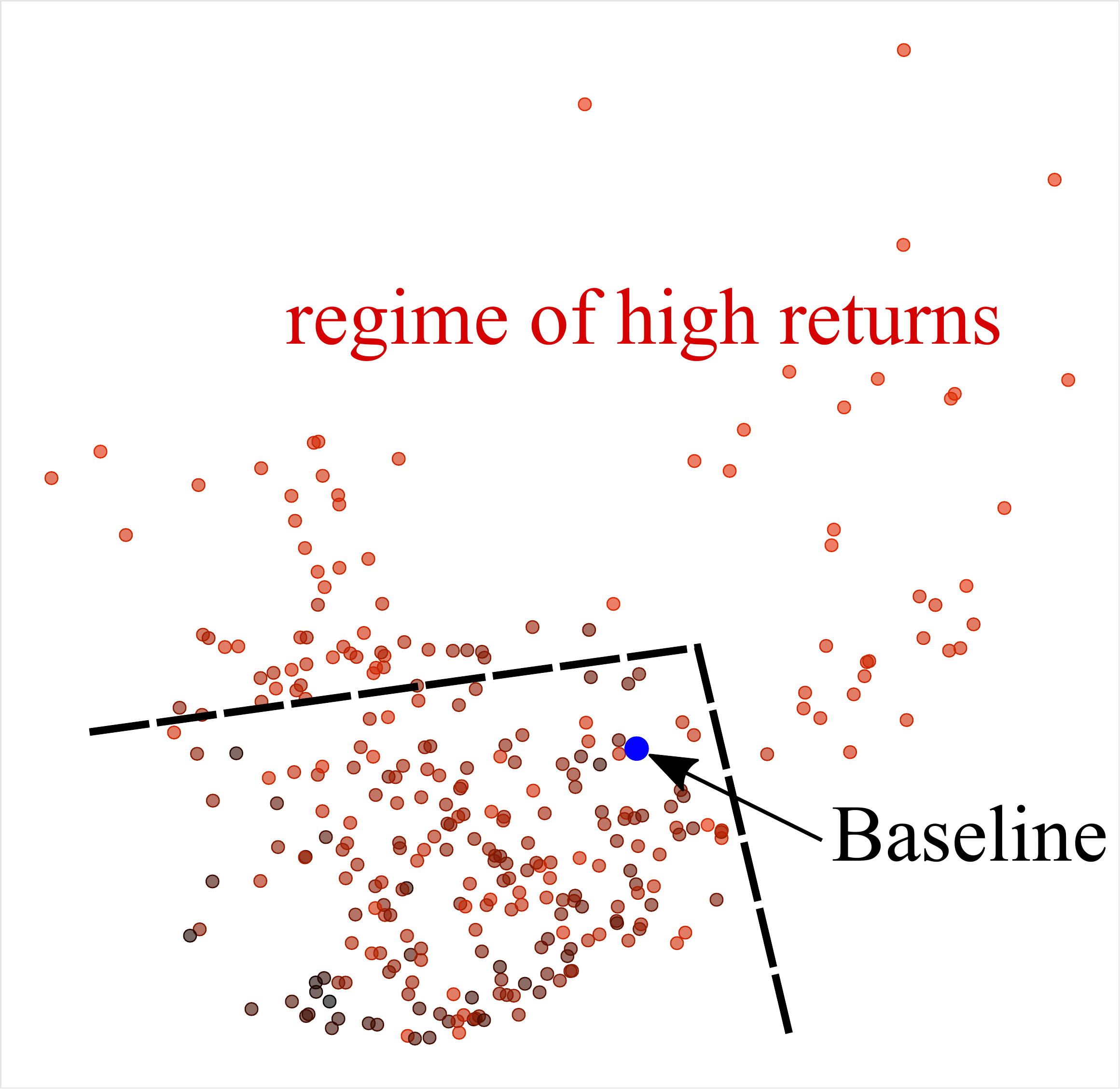}
        \subcaption{First layer, PSU}
    \end{subfigure} 
    \hfill
    \begin{subfigure}{.32\textwidth}
        \centering
        \includegraphics[width=\textwidth]{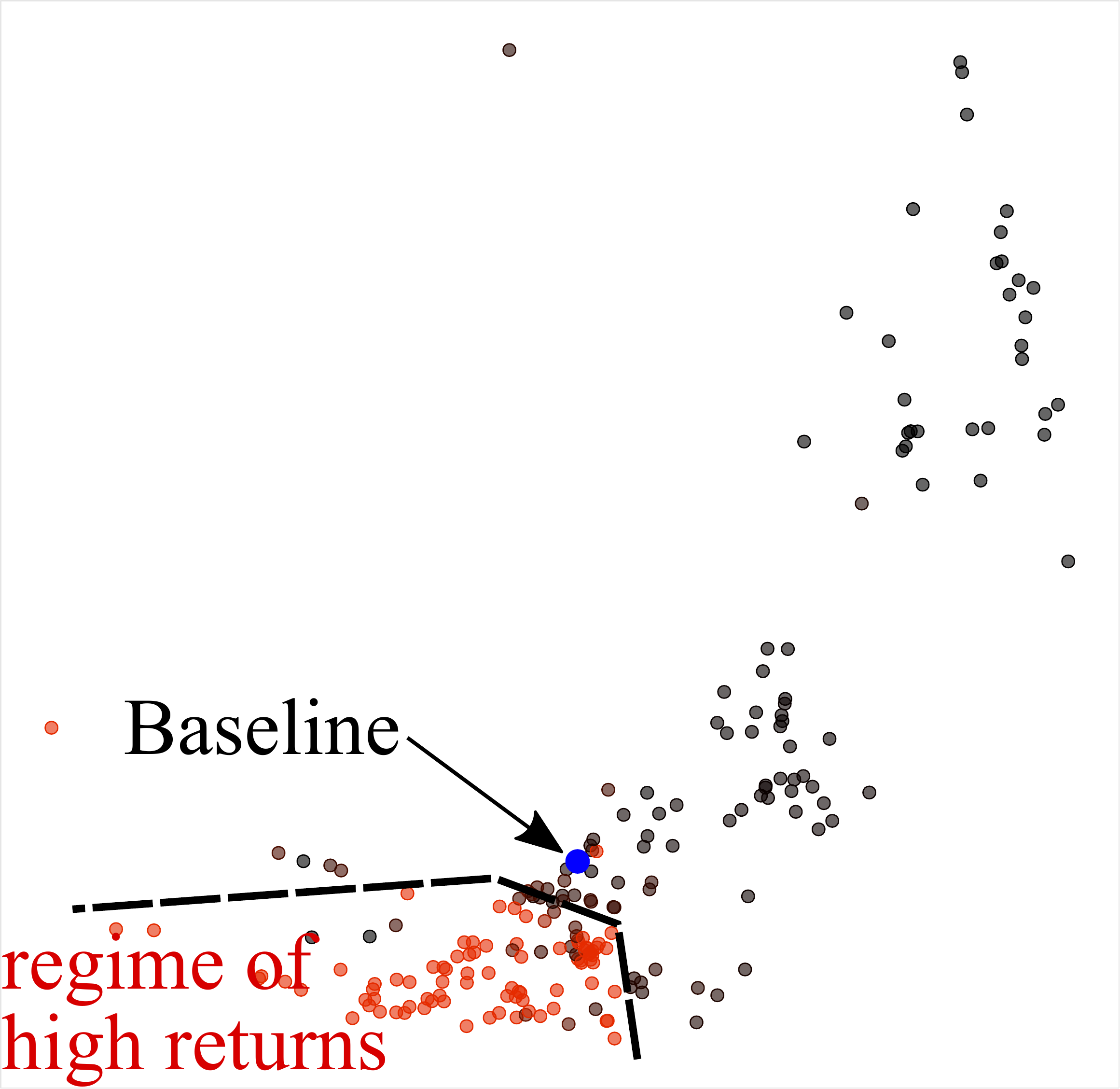}
        \subcaption{Second layer, CPB}
    \end{subfigure}
    \caption{Scatterplot showing the mean (x-axis) and variance (y-axis) of the correlation coefficients for all ablations of the specified layer.}
    \label{fig:activation_actions_correlations_mean_var_scatter_all}
\end{figure}

To further test the validity of the hypothesis across different sizes of ablations and across the three tasks, Figure \ref{fig:activation_actions_correlations_mean_var_scatter_all} summarizes the effects of all ablations ($5\%$ to $90\%$) on the return and the dependency on the characteristics of the correlation patterns.
Analogously to figure \ref{fig:invertedpendulumswingup_linear1_activation_action_correlations_05_scatter}, the x- and y-axis show the mean and the variance of the correlation patterns. For the CPSU task, the highest return is generally achieved for patterns with a low variance as ablations leading to larger variances show a decreased return. This suggests that the CPSU task requires single units to be generically involved in the control task and not to specialize too strongly on specific controls. On the contrary for the PSU task, higher returns are generally achieved for patterns with a high mean and high variance, suggesting a further refinement of our hypothesis with respect to task specific characteristics. Interestingly, ablations that increase both values beyond the baseline lead to even higher returns while patterns with low values lead to low returns. This suggests that the ability to swing-up the pendulum requires the units to contribute to the control in a very specific rather than generic way. Consistently, a very clear picture emerges for the CPB task, where no swing-up is required and only patterns with low values for mean and variance result in high returns, verifying our initial hypothesis. In combination with the CPSU task, this suggests that the ability to control the moving cart requires a generic involvement of single units in the control task rather than specific roles.

\subsection{Impact of Ablations on Layer Activation}
Although the correlation patterns provide some insights on how the agent acts, they do not capture the temporal evolvement of the learned representations and do not answer questions with respect to such evolvements, e.g. at what point during the episode does the agent fail? When does it diverge from the baseline behavior and in what way? Does the agent go through different behavioral stages during an episode and can these stages be linked to specific patterns in the the learned representation? In order to answer these questions, we characterize the learned representations by embedding the layer activations recorded during an episode (cf. \ref{sec:method_patterns}) and compare the representations of the baseline to the representations resulting from the ablations. 

\begin{figure}[t!]
\centering
\tabskip=0pt
\valign{#\cr
  \hbox{%
    \begin{subfigure}{.66\textwidth}
    \centering
    \tikz\draw[cw-red,fill=cw-red] (0,0) circle (.4ex); Ablated \hspace{1.5cm}
    \tikz\draw[cw-blue,fill=cw-blue] (0,0) circle (.4ex); Baseline
    \\ \vspace{0.2 cm}
    \includegraphics[width=\textwidth]{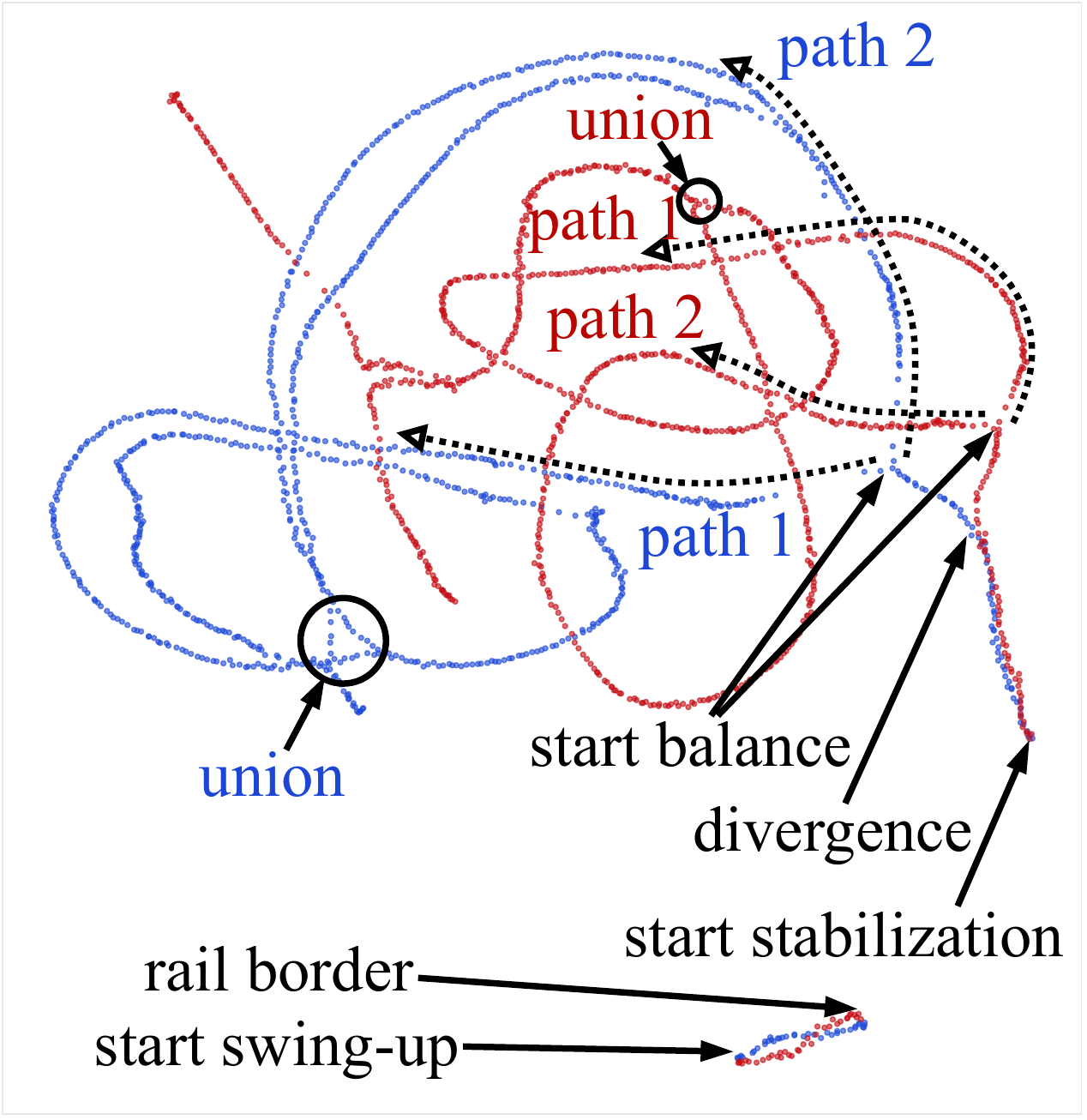}
    \caption{Ablation of units 20 to 39 ($5\%$) in the first layer.}
    \end{subfigure}%
  }\cr
  \noalign{\hfill}
  \hbox{%
    \begin{subfigure}{.33\textwidth}
    \centering
    \includegraphics[width=\textwidth]{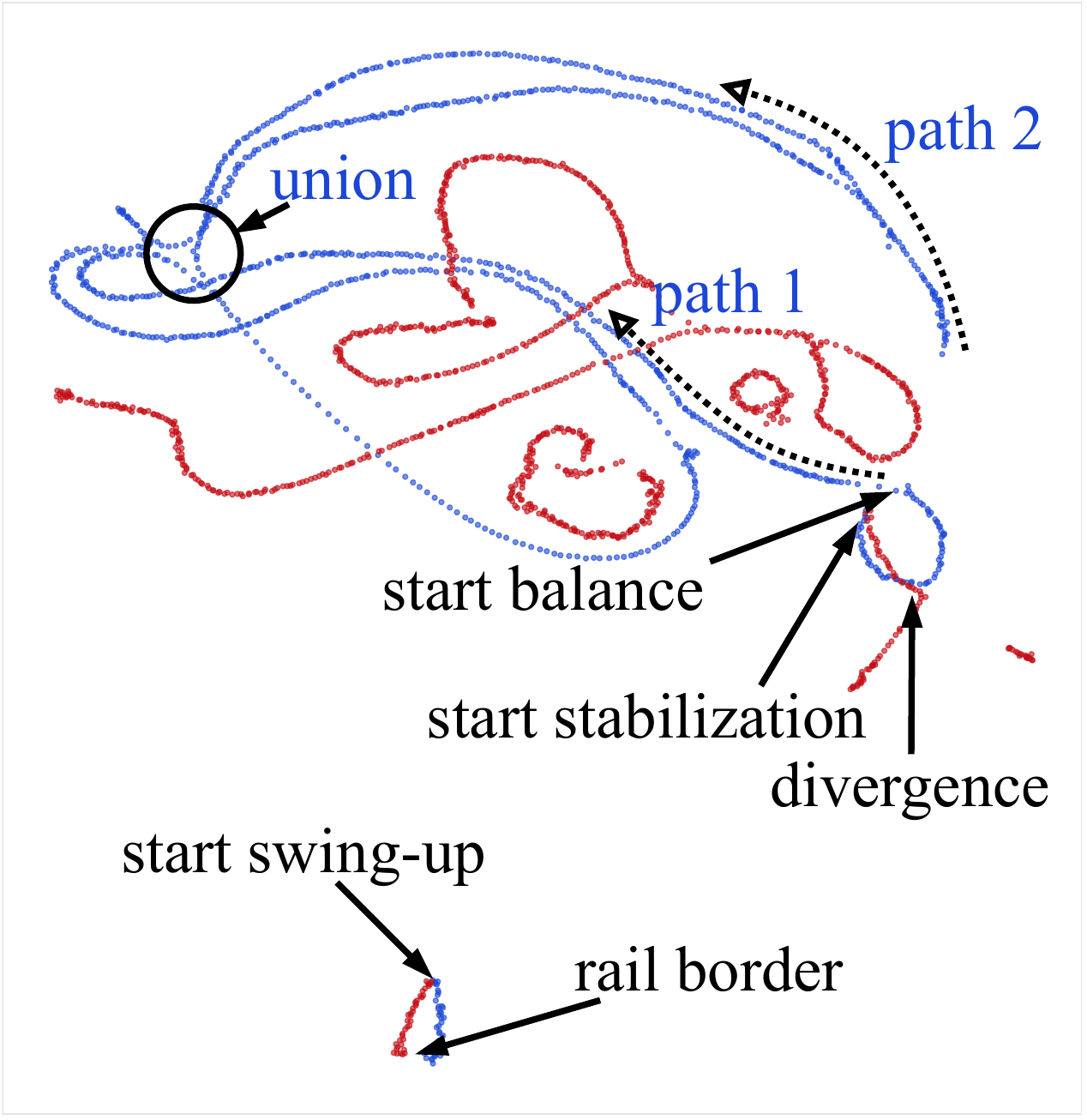}
    \caption{Units 110-149 ($10\%$), layer 1.}
    \end{subfigure}%
  }\vfill
  \hbox{%
    \begin{subfigure}{.33\textwidth}
    \centering
    \includegraphics[width=\textwidth]{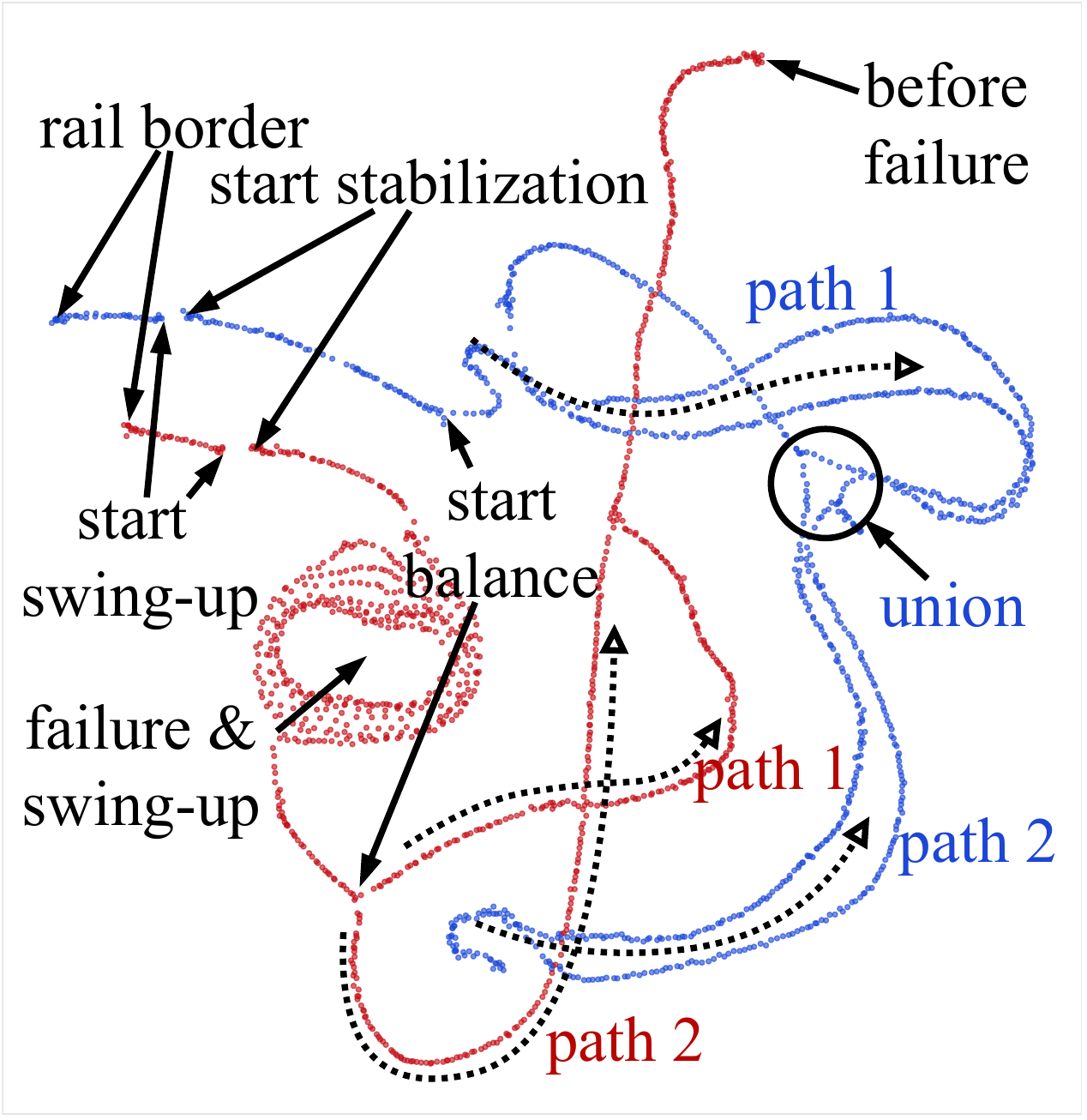}
    \caption{Units 260-289 ($10\%$), layer 2.}
    \end{subfigure}%
  }\cr
}
\caption{Comparison of the temporal evolvement of layer activations between the baseline and three exemplary ablation cases for the CPSU task.}
\label{fig:link_activation_behavior}
\end{figure}

Figure \ref{fig:link_activation_behavior} shows this comparison for three exemplary ablation cases for the CPSU task. Each scatter plot contains 1000 blue and 1000 red points corresponding to the layer activation for each time step during an episode for the baseline and the ablation case, respectively. Note that even though the baselines in (a) and (b) show the exact same values, they are embedded slightly differently as the embeddings were calculated separately for all cases. The three cases correspond to ablations, which had no effect on the agent's capability to solve the task (Figure \ref{fig:link_activation_behavior}a) or which lead to only half the return of the baseline (Figure \ref{fig:link_activation_behavior}b and c).
Figure \ref{fig:link_activation_behavior}a shows the evolvement of the layer activation during an episode for the healthy and the damaged agent and how the different behavioral stages of the episode are linked to different sections of this evolvement. Both, the healthy and the damaged agent, start with moving the cart to the side, accelerating the pendulum to swing it up. After the initial swing-up (upon reaching the rail border), the agent is required to compensate for the excess momentum of the pole via corresponding cart movement to stabilize its upright position. This change in behavior results in a jump in the activation space from the initial activation path that corresponds to the initial swing-up behavior to another path that corresponds to the stabilization behavior. The difference in activations is likely due to the movement of the cart into the opposite direction upon reaching the rail border. Following the successful stabilization, the agent is required to balance the pole by rapidly switching directions of the cart to maintain an upright pole position. Interestingly, this behavior is represented in the activation space by two paths, along which the layer activation progresses as the agent acts throughout the episode. The layer activation repeatedly switches between these two paths suggesting that the network constantly changes between two distinct activation states corresponding to the balancing act of the pole. At some point during the episode, these two paths merge together (union) as the balancing act leads to an almost static position of the cart and the pole. However, from a mechanical perspective, this constitutes an unstable equilibrium point for the pole, where small perturbations of the pole's angular position result in its downfall triggering a renewed balancing act that is resembled by a renewed separation of the merged paths. This observations suggests that the convergence of the actor's activation towards a single final activation state is not sufficient to solve the task. Rather, a stable and continuous transition between two distinct activation states is necessary to sufficiently represent the balancing act. This observation seems somewhat surprising considering the weak correlations of SUA to the actor's chosen actions throughout an episode (cf. \ref{sec:results_SUA}). Although the SUA does not correlate strongly with the network's executed actions, their combined activations lead to two distinct activation states of the network, each of them corresponding to the movement of the cart in either one of the two possible directions during the balancing act. This suggests that single units do not contribute individually to the control task, but rather as part of a larger conglomerate of units that constitute the two different activation states.

Figure \ref{fig:link_activation_behavior}b shows an ablation case, for which the agent fails to balance the pole continuously after the initial swing-up and drops it after a short period of holding it in the upright position, reattempting the swing-up and balancing act. The layer activation diverges slightly from the baseline right from the start of the swing-up and further diverges completely after a short period of the stabilization phase. Consequently, due to this divergence, the layer activation of the damaged agent does not show the emergence of two distinct paths connected to the balancing act as the agent never succeeds in stabilizing the pole compensating its excess momentum after the initial swing-up. 
Interestingly, the existence of two distinct activation states is not exclusive to the actor's first layer but also apparent in its second layer. Figure \ref{fig:link_activation_behavior}c shows an ablation case in layer two, in which the failure of the agent is caused by a drop of the pole after the initial swing-up and a short period of balancing, causing the pole to rotate at high speed until the end of the episode. The blue points resemble a similar pattern of the second layer's activation compared to the first layer including the divergence of the activation along two distinct paths, the attempt to merge these paths and the renewed separation. The failure of the agent, i.e. the continuous rotation of the pole at high speed, is visible in the activation space by the circularly arranged red points, from which the agent is not able to recover back onto the stabilization path and the both connected paths corresponding to the balancing act.

\section{Conclusions \& Future Work}
In this paper, we conducted an empirical study to understand how a DRL agent acts based on characterizing the learned representations of its policy network. We shed some light on the role of single units for the control task and found that despite the absence of a strong correlation between their activations and the actor's chosen actions throughout an episode, agents, that solve their tasks successfully, show task specific patterns of weakly correlated SUA that get distorted by network ablations leading to low returns. The importance of these patterns for a successful solution of the control task suggests that the careful interplay between single units with respect to the executed policy is essential rather than their sole and isolated behavior. However, we have only scratched the surface of how such patterns of joint activations can be characterized. In our future work, we plan to systematically investigate the role of functional neuron populations and their involvement in solving a given control task. Specifically, we plan to investigate the activation of sub-populations of neurons aiming to uncover if there is a link between their activations and the emergent agent behavior.

We further investigated the temporal evolvement of the actor's layer activations during an episode and showed that, in case of the CPSU task, the consecutive steps executed during the episode to solve the task are precisely represented by the policy network and mapped onto its layer activations. We fruther showed that this mapping is essential for solving the task as its distortion as a result of network ablations leads to low returns and failed attempts to solve the task. The arrangement of the consecutive points in the embedded activation space revealed that the agent runs along specific paths in its activation space and that diverging from this path is fatal for its task performance. The most striking observation of these paths is given by the fact that the actor's layer activations can be very different for very similar states. We naively expected that the layer activation would converge to a single specific activation vector just as the consecutive states to be processed by the network become more and more similar to each other as the pole is balanced. However, we found that this is not the case, suggesting that the learned representations may contain some information that is encoded in the temporal dimension on which the states are ordered, i.e. that the same state evokes a different activation of the network depending on when it is presented to the network. In our future work, we plan to investigate how these distinct activation patterns evolve during training, aiming to answer the question, whether the different behaviors are learned hierarchically, i.e. in a specific order, or whether they emerge collectively.

Considering that our study was limited to a single agent solving three distinct control tasks, the universality of our results is strongly limited and their implications for other networks and tasks is not clear. We plan to address this issue by transferring our study design to a larger number of different networks and control tasks aiming to establish a scientific standard for the falsifiability of empirical studies conducted in the field of artificial neural networks. Ultimately, we aim to pave the way towards a new perspective of neuroscience inspired empirical studies on artificial neural networks to exploit them as a test bed for neuroscientific research. Uncovering parallels between the structure and organization of represented knowledge in artificial and biological systems opens up measures and possibilities for initial large scale studies in artificial systems before transferring them to biological systems. Specifically, this addresses the issue of reproducibility, which, despite modern experimental methods, is one of the most critical issues in modern neuroscience, stemming from the large differences between brains and the commonly small sample sizes in neuroscientific studies.
%
%
%
\bibliographystyle{ieeetr}
\bibliography{main}
\end{document}